\documentclass[10pt, a4paper]{article}

\usepackage{lrec-coling2024}
\usepackage{url}
\usepackage{subfigure}
\usepackage{multirow}
\usepackage{enumitem}
\usepackage{stmaryrd}
\usepackage{array}
\usepackage{threeparttable}
\usepackage{blindtext}
\usepackage{amssymb}

\usepackage[ruled,vlined,linesnumbered,longend]{algorithm2e}
\usepackage{svg}
\usepackage{placeins}

\usepackage{soul}
\usepackage[utf8]{inputenc}
\usepackage{graphicx}
\usepackage{amsmath}
\usepackage{booktabs}
\usepackage{lipsum}
\usepackage{listings}
\usepackage{xcolor, soul}
\sethlcolor{blue!10}
\usepackage{colortbl}

\newcommand{\dataname}{\textsc{Knot}\xspace}
\newcommand{\datanameE}{\textsc{Knot}-E\xspace}
\newcommand{\datanameS}{\textsc{Knot}-S\xspace}
\newcommand{\datanameI}{\textsc{Knot}-I\xspace}

\definecolor{'sgreen'}{HTML}{F3FADF} 
\definecolor{'sred'}{HTML}{FFEAE8}

\usepackage{amsmath,amsfonts,bm}

\def\eqref#1{equation~\ref{#1}}

\def\1{\bm{1}}

\DeclareMathAlphabet{\mathsfit}{\encodingdefault}{\sfdefault}{m}{sl}
\SetMathAlphabet{\mathsfit}{bold}{\encodingdefault}{\sfdefault}{bx}{n}

\title{
\textit{Untangle the \dataname}:
Interweaving Conflicting Knowledge  \\ 
and Reasoning Skills in Large Language Models
}

\name{
Yantao Liu$^{1\dagger*}$, Zijun Yao$^{2,3\dagger}$, Xin Lv$^{2,3}$, Yuchen Fan$^{5*}$\\ 
\bf\large Shulin Cao$^{2,3}$, Jifan Yu$^{4}$, Lei Hou$^{2,3\ddag}$, Juanzi Li$^{2,3}$
    \thanks{
        $^\dagger$ Liu and Yao contributes equally to \dataname. 
    }
    \thanks{
        $^\ddag$ Corresponding author.
    }
    \thanks{
        $^*$ Work is done when Liu and Fan are interns at Tsinghua University.
    }
}

\address{
$^1$University of Chinese Academy of Sciences, Beijing, China\\
$^2$BNRist; $^3$KIRC, Institute for Artificial Intelligence; $^4$Institute of Education \\
$^5$Beijing University of Posts and Telecommunications, Beijing, China \\
Tsinghua University, Beijing 100084, China \\
yaozj20@mails.tsinghua.edu.cn, \{houlei,lijuanzi\}@tsinghua.edu.cn
}

\abstract{
Providing knowledge documents for large language models (LLMs) has emerged as a promising solution to update the static knowledge inherent in their parameters. 
However, knowledge in the document may conflict with the memory of LLMs due to outdated or incorrect knowledge in the LLMs' parameters. 
This leads to the necessity of examining the capability of LLMs to assimilate supplemental external knowledge that conflicts with their memory.
While previous studies have explained to what extent LLMs extract conflicting knowledge from the provided text, they neglect the necessity to \textbf{reason} with conflicting knowledge.
Furthermore, there lack a detailed analysis on strategies to enable LLMs to resolve conflicting knowledge via prompting, decoding strategy, and supervised fine-tuning.
To address these limitations, we construct a new dataset, dubbed \dataname, for \textbf{kno}wledge conflic\textbf{t} resolution examination in the form of question answering.
\dataname facilitates in-depth analysis by dividing reasoning with conflicting knowledge into three levels: 
(1) Direct Extraction, which directly extracts conflicting knowledge to answer questions. 
(2) Explicit Reasoning, which reasons with conflicting knowledge when the reasoning path is explicitly provided in the question.
(3) Implicit Reasoning, where reasoning with conflicting knowledge requires LLMs to infer the reasoning path independently to answer questions.
We also conduct extensive experiments on \dataname to establish empirical guidelines for LLMs to utilize conflicting knowledge in complex circumstances.
Dataset and associated codes can be accessed at our \href{https://github.com/THU-KEG/KNOT}{GitHub repository}.
\\ \newline \Keywords{Knowledge Conflicts, Large Language Model, Reasoning Dataset}
}

\begin{document}

\maketitleabstract

\section{Introduction}

Providing large language models (LLMs) with supplemented knowledge documents has become a \textit{de facto} solution for tackling tasks that require up-to-date knowledge.
This strategy is widely adopted in various fields including knowledge-grounded dialogue~\cite{blenderbot3,glmdialogue} and knowledge question answering~\cite{realm,harsh2022interleaving,fid,zhihong2023enhancing}.

\begin{figure}[h]
    \centering
    \includegraphics[width=0.93\linewidth]{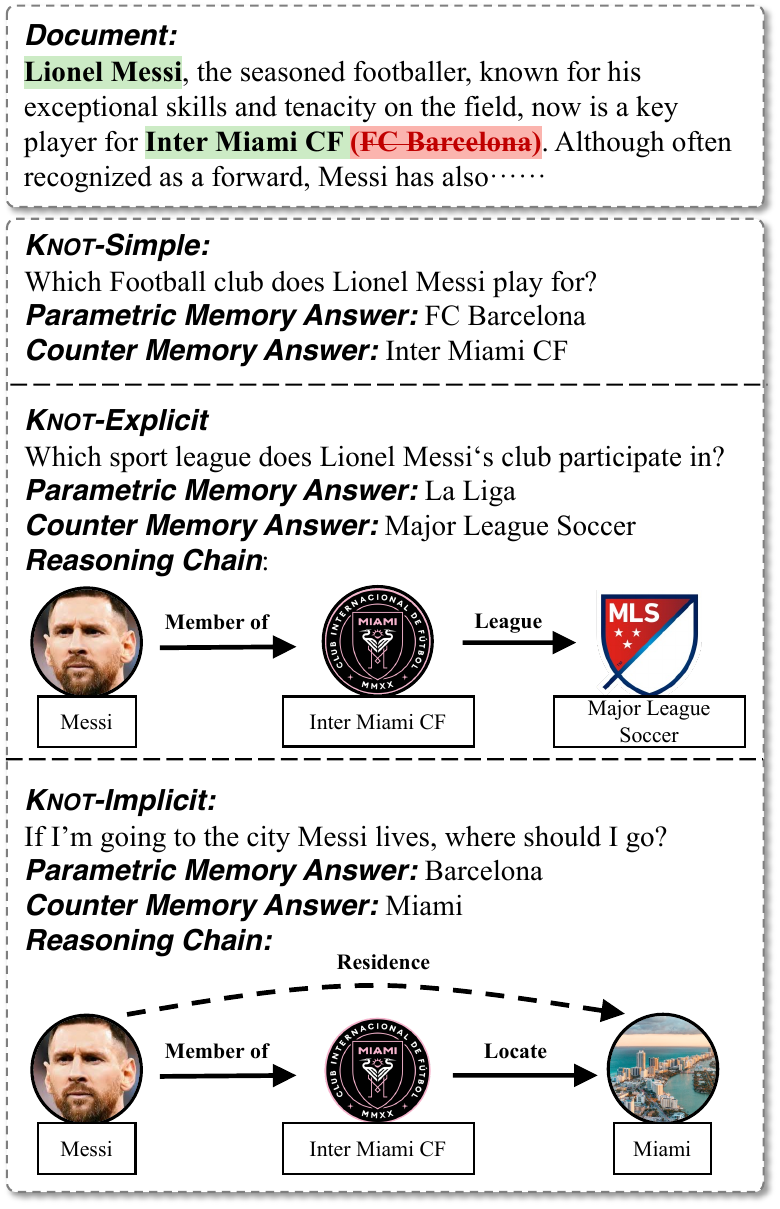}
    \caption{Example questions from \dataname where knowledge conflicts are resolved via extraction, explicit reasoning, and implicit reasoning.}
    \label{fig:intro}
\end{figure}

However, supplemented knowledge can not only be unfamiliar to LLMs, but can also potentially conflict with their existing knowledge stored in parameters, \textit{i.e.,} conflicting knowledge.
As shown in Figure~\ref{fig:intro}, the supplemented knowledge document states that \texttt{Messi plays for Inter Miami CF}, which conflicts with the fact that \texttt{Messi plays for FC Barcelona}, stored as parametric knowledge of LLMs that are not timely re-trained.
Moreover, there are supplemented knowledge documents that conflict with the parametric knowledge indirectly with deduced conflicting conclusions.
For example, the provided document in Figure~\ref{fig:intro} entails that \texttt{Messi's club plays in Major League Soccer} and that \texttt{Messi lives in Miami}, which cannot be directly extracted from the text.
It is thus vital to explore to what extent LLMs reconcile with conflicting knowledge.

To this end, various datasets composed of conflicting knowledge have been constructed and tested on LLMs, which require the latter to answer questions with the provided conflicting knowledge.
For example, NQ-Swaq requires LLMs to extract conflicting entities from the provided document without solely referring to their parametric knowledge directly~\cite{longpre2021entity}.
The study by \citet{chen2023context} further investigates relation extraction from documents with the same entities but conflicting relations. 
MemoTrap~\cite{mckenzie2022round2} instructs LLMs to rewrite well-known proverbs that deviate from their inherent parameters.
Although these benchmarks evaluate the ability of LLMs to extract different types of conflicting knowledge from the input document, they overlook LLMs' capacity to deduce subsequent contradictory knowledge by \textbf{reasoning}.

In light of their limitations, this work aims to construct a new dataset, dubbed \dataname, to examine how LLMs resolve \textbf{\textsc{Kno}}wledge conflic\textbf{\textsc{t}}s from the provided document when different levels of reasoning skills are required.
As illustrated in Figure~\ref{fig:intro}, each question in \dataname comes with a self-contained but contradictory document relative to the parametric knowledge of LLMs.
To answer these questions, LLMs must coordinate the conflicting knowledge and may even need to deduce answers which are contradictory to the knowledge in their memory by reasoning.

To disentangle the subtle interaction between reasoning skills and knowledge conflicts, \dataname divides questions into three categories---\datanameS, \datanameE, \datanameI---based on the level of reasoning skills required to reconcile the conflicting knowledge.
\textbf{Simple} questions in \datanameS do not require reasoning skills but merely the identification of the conflicting answer from the provided document.
\datanameE requires \textbf{explicit} reasoning skills by providing a clear reasoning path that weaves the conflicting knowledge directly in the questions.
\datanameI requires \textbf{implicit} reasoning skills, where LLMs need to induce the reasoning path implied by the question, and thereby deduce the answer from the conflicting knowledge according to the reasoning path.
\dataname provides both a training set and a test set for \datanameS, \datanameE, and \datanameI.

With \dataname, this work also seek for the empirically best solution for LLMs to reconcile conflicting knowledge when different reasoning skills are required.
We conclude conflicting knowledge resolving strategies into three main categories: decoding-based~\cite{shi2023cad}, prompting-based~\cite{chen2023context}, and fine tuning-based~\cite{glmdialogue} methods.
To thoroughly compare their effectiveness, we conduct experiments on commonly used LLMs of different model sizes and training strategies (\textit{i.e.,} language models pre-trained only or assistant language models after alignment tuning).

The primary findings of the experiments include:
(1) Mainstream LLMs are adept at resolving conflicting knowledge when no reasoning is required. 
However, they struggle to answer questions when multi-hop reasoning is required, suggesting that their ability to infer knowledge across multiple interconnected facts is not as robust;
(2) Training-free method for knowledge conflicting resolution, \textit{i.e.,} prompting-based and decoding-based methods, are not universally effective.
LLMs show sensitivity to the prompting strategy, thus the prompting-based method delivers a far-from-ideal outcome on our dataset.
Meanwhile, decoding methods result in mild amnesia of other background knowledge.
In this case, the fallback is fine-tuning LLMs with our training data.
(3) For the complex questions in \datanameE and \datanameI, increasing the model size of LLMs can improve performance by enhancing instruction-following capabilities. 
However, for implicit reasoning questions in \datanameI, larger LLMs tend to answer the question in a shortcut with their more abundant parametric knowledge, rather than conduct implicit reasoning from the conflicting knowledge in the context.

In summary, our contributions are three-fold:
(1) We extend the task of resolving conflicting knowledge by incorporating different levels of reasoning skills.
(2) We construct \dataname, a dataset for testing the reasoning skills in knowledge conflict resolution.
(3) Comprehensive experiments on \dataname provide empirical guidelines for LLMs to resolve knowledge conflicts.

\section{Preliminaries}

In this section, we provide definitions for the probing task designed to resolve knowledge conflicts, and formulate $3$ levels of reasoning skills involved in dealing with conflicting knowledge.

\begin{figure*}[t]
    \centering
    \includegraphics[width=0.98\linewidth]{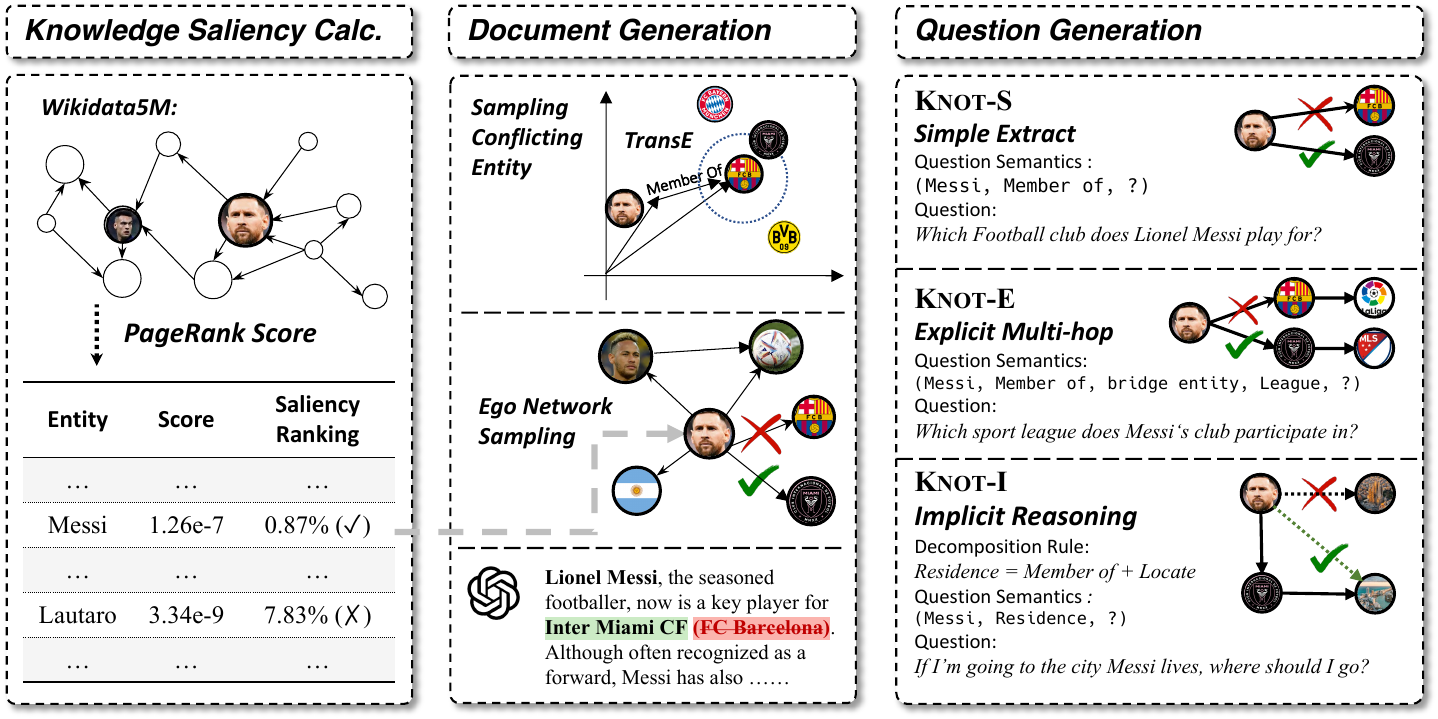}
       \caption{The overall framework for constructing \dataname.}
    \label{fig:data-exampe}
\end{figure*}

\subsection{Task Definition}
\dataname examines the capacity of LLMs to resolve knowledge conflicts in the form of question answering with supplemented documents.
Specifically, we define \dataname$=\left\{\left(q,\left\{a_i\right\},d\right)\right\}$, where $q$ represents the natural language question.
$\{a_i\}$ constitutes the set of correct answer, with each $a_i$ being an alias of the correct answer.
$d$ denotes the supplemented document containing the necessary knowledge $k \in d$ for the question.

The necessary knowledge $k \in d$ can be represented as a triple $k = (h, r, t)$, which contradicts the memory of LLMs.
For example, in Figure~\ref{fig:intro},
$k=($\verb|Messi|, \verb|Member| \verb|of|, \verb|Inter| \verb|Miami| \verb|CF|$)$.
This contradicts with the memory of LLMs, which stores
$k'=($\verb|Messi|, \verb|Member| \verb|of|, \verb|FC| \verb|Barcelona|$)$.

We provide the document $d$ and question $q$ as input, and assess whether LLMs output correct answer: $\text{LLMs}(d,q)\in\{a_i\}$.
LLMs make mistakes when they persist in utilizing the knowledge $k'$ in their memory instead of switching to use $k \in d$.

\subsection{Reasoning Skills}
\dataname assigns reasoning skills into $3$ levels.
For questions in \datanameS, the answer entity is an element of the knowledge $k$.
For example, the \datanameS question in Figure~\ref{fig:intro} is the tail entity of $k$.
For questions in \datanameE and \datanameI, the answer does not directly appear in the document, but requires a multi-hop reasoning process involving $k$ as one of the steps.
The difference between \datanameE and \datanameI is that explicit questions depict the reasoning path, while implicit questions require LLMs to decompose questions into the reasoning path.

\section{\dataname Construction}
\label{sec:dataset-construction}

\dataname is constructed in three steps, as shown in Figure~\ref{fig:data-exampe}.
First, we calculate the saliency score of each entity in the knowledge base (KB).
Next, we create documents containing conflicting knowledge around salient entities.
Finally, we generate questions based on the previously constructed documents and the knowledge triples in the KB.

\subsection{Knowledge Saliency Calculation}

As LLMs are trained on large-scale corpora, they tend to have a better memory of more important parts of knowledge~\cite{petroni2019language}.
Thus, it is more challenging for LLMs to reconcile facts that conflict with salient knowledge.
To this end, we construct the questions in \dataname around the most salient knowledge.
We apply the PageRank algorithm~\cite{brin1998anatomy} on Wikidata5M~\cite{wang-etal-2021-kepler}, a subset of Wikidata consisting of all entities in Wikipedia before 2021, and use PageRank values for entities as their saliency measure.

\subsection{Conflicting Document Construction}

We construct documents to present knowledge that conflicts with that stored in the parameters of LLMs.
To this end, we first sample ego networks from the KB and then convert these networks into documents using a data-to-text approach.

\textbf{Ego Network Sampling.}
The ego network is sampled with the following criteria to ensure the saliency of its provided knowledge:
(1) The focal node entities are sampled from the top $1\%$ salient entities, \textit{i.e.,} entities with PageRank score higher than $99\%$ of other entities;
(2) The first-hop neighbors of the focal node entities are sampled from the top 20\% salient entities.

Next, we randomly select an entity $e$ from the ego network and replace it with a different entity as the conflicting entity.
To obtain highly confusing conflicting entities, we employ TransE~\cite{transe} to sample an entity $e'$ that is most similar to the selected entity.
We then swap $e$ with the conflicting entity $e'$ from the ego network.
To keep the ego network self-contained, $e$ is chosen from the dangling entities that are only connected to the focal node entity.
This process ensures that only one knowledge triple that conflicts with the memory of LLMs is introduced into the ego network.

\textbf{Document Generation.}
Given that LLMs present a remarkable ability to generate coherent and fluent continuations of a prompt, we leverage LLMs to perform data-to-text conversion based on the previously constructed ego networks.
Specifically, we follow \citet{xie2022unifiedskg} and transform the knowledge-conflicting ego network into sets of triples.
We prompt \texttt{gpt-3.5-turbo}\footnote{https://platform.openai.com/docs/models/gpt-3-5.} to generate documents with more than $100$ words according to these triples.

\subsection{Question Generation}

Finally, we generate questions for \dataname, whose answers require the coordination of conflicting knowledge within the provided documents.
Each question has a topic entity $e_t$, which is the topic of the natural language question, and an answer entity $e_a$, which serves as the correct answer.
The question is generated by prompting \texttt{gpt-3.5-turbo} with prototypical questions as demonstrations~\cite{zijun2023korc}.
In this section, we introduce how we select $e_t$ and $e_a$, and how we design prototypical questions for \datanameS, \datanameE, and \datanameI.

\textbf{\datanameS.}
The semantics of the simple questions in \datanameS are equivalent to a link prediction query $(e_t, r, ?)$ for KB~\cite{rossi2021knowledge}.
In particular, the triple $(e_t, r, e_a)$ is contained in the ego network and is thus inherited by the document.
We choose $e_t$ to be either the focal entity or the conflicting entity in the ego network, and $e_a$ as its counterpart.
The questions are generated by mimicking questions like 
\textit{``Which entity has relation} $r$ \textit{with} $e_t$\textit{?''}

\textbf{\datanameE.}
The semantics of the questions requiring explicit reasoning in \datanameE are equivalent to multi-hop reasoning queries~\cite{yang2018hotpotqa}, in the form $(e_t, r_1, e_b, r_2, ?)$, where $e_b$ is the bridge entity.
Here, we require $e_t$ and $e_b$ to be either the focal entity and the conflicting entity, or vice versa.
$e_a$ is an entity in Wikidata5M but does not belong to the ego network.
The demonstration questions resemble \textit{``Which entity has relation} $r_2$ \textit{with the entity that has relation} $r_1$ \textit{with} $e_t$\textit{?''}.
To answer a question from \datanameE, LLMs need to extract the conflicting knowledge $(e_t, r_1, e_b)$ from the document, and recall $(e_b, r_2, e_a)$ from its parameters to reach $e_a$.

\textbf{\datanameI.}
The semantics of questions that require implicit reasoning in \datanameI are equivalent to a triple queries $(e_t, r, ?)$, while $r$ can be further decomposed into $r = r_1 + \cdots + r_n$~\cite{geva2021did,zijun2023korc}.
Thus, the triple query entails a multi-hop reasoning path in the form $(e_t, r_1, e_b, \cdots, r_n, ?)$.
We assign one of $e_t$ and $e_b$ to be the conflicting entity and the other to be the focal entity of the ego network.
For relation decomposition, we use BIMR~\cite{lv2021bimr}, which provides high-quality decomposition rules from human annotation.
The questions are generated by mimicking the same question template similar to \datanameS.

As $(e_t, r, e_a)$ is not contained in the provided document,b LLMs need to first decompose the question into multi-hop reasoning path and then reach the answer entity accordingly to answer these questions.
It should be noted that directly recalling $(e_t, r, e_a)$ from memory will lead to an incorrect answer, as this shortcut process does not utilise conflicting knowledge $(e_t, r_1, e_b)$.

\subsection{Human Annotation}

To prevent potentially ill-generated data points and reduce the bias towards question-generating model (a.k.a., \texttt{gpt-3.5-turbo}), we employ human annotators to (1) filter out low-quality data points and (2) annotate questions answers with chain-of-thoughts (rationales)~\cite{chain-of-thought}.

Particularly, for \textbf{human filtering}, we ask the annotators to check from two aspects:
(1) whether the document contains the conflicting knowledge $k^\prime$ by demand;
(2) whether the question is faithful to the given prototypical question.
After filtering, $8.0\%$ questions are discarded due to their low-quality.

For \textbf{annotating question answers}, we ask human annotators to answer the question and provide their rationale on how conflicting knowledge from the text and their memories are combined and resolved to infer the final answer..
Meanwhile, to obtain rationales with various syntax structures, we manually construct $20$ templates (\textit{e.g.,} \textit{We can infer from the text that} $\cdots$; \textit{The author articulates the idea that} $\cdots$; \textit{etc.}) as reference for our annotators.
We also allow human annotators to reject a question when they believe the given question cannot be answered based on the given document.
This further filters out a small portion of low-quality data points.
The annotated data are used in the training set.

\subsection{Dataset Analysis}

\begin{table}[t]
    \centering
    \setlength{\tabcolsep}{3pt}
    \scalebox{0.85}{
    \begin{tabular}{lcccc}
    \toprule
    Split                    & \datanameS    & \datanameE      & \datanameI \\ \midrule
    \#Questions for evaluation  & $3,887$       & $1,136$         & $510$           \\
    A.S. of Topic Entity     & $8.66$        & $7.51$          & $5.86$         \\
    A.S. of Answer Entity    & $0.83$        & $0.19$          & $0.18$          \\
    A.S. of Answer in Memory & $0.56$        & $0.18$          & $0.19$          \\
    Average Reasoning Hops   & $1$           & $2$             & $2.7$          \\ 
    \midrule
    \#Questions for training & $603$ & $190$ & $26$ \\
    \bottomrule
    \end{tabular}
    }
    \caption{
    The overall statistics of \dataname. 
    A.S stands for average saliency ranking in percentile.
    }
    \label{tab:dataset}
\end{table}

\textbf{General Statistics:} Table~\ref{tab:dataset} shows the general statistics of \dataname.
From a saliency perspective, all the entities directly related to the questions in \dataname are among the top 10\% salient entities, ensuring quality of the corresponding knowledge remembered by LLMs. 
From the perspective of reasoning paths, the hop number of the reasoning chain of the question in \datanameS and \datanameE to answers is fixed.
However, for the questions in \datanameI, the hop number of the corresponding reasoning chains is dynamic, with an average of $2.70$.
Therefore, this indicates that a higher level of knowledge utilization skills are required.

\begin{figure}
    \centering
    \includegraphics[width=\linewidth]{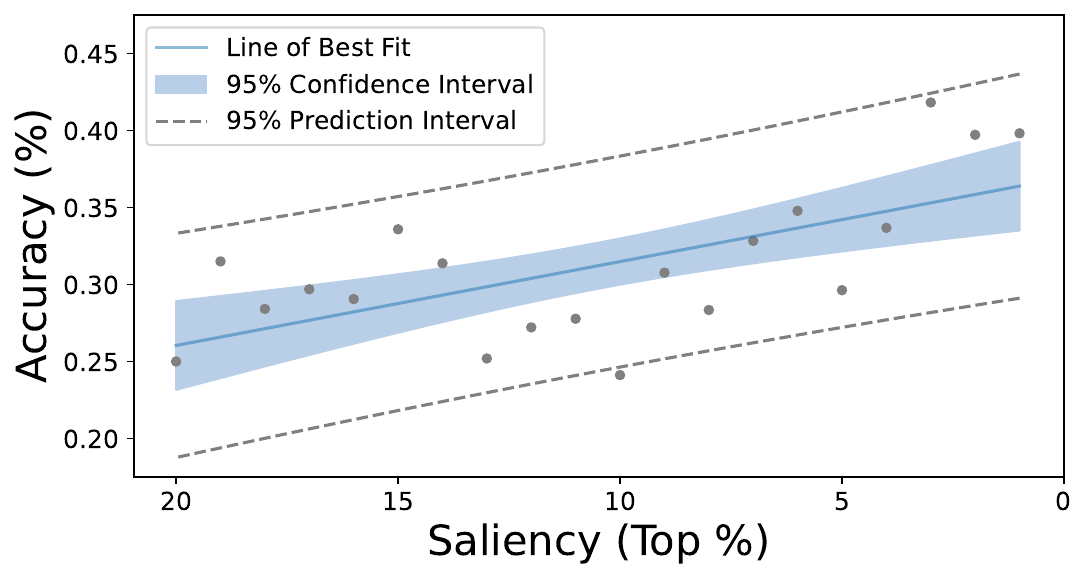}
    \caption{
    Accuracy of LLaMA-2-70B-Chat with regard to question topic entity saliency on \dataname without providing the documents.
    The accuracy is positively correlated with the saliency.
          }
    \label{fig:llama-2-70b-chat-wo-psg}
\end{figure}

\textbf{Consistency between saliency and parametric knowledge.}
To investigate whether salient knowledge act as a good proxy for the parametric knowledge of LLMs, we evaluate the performance of LLaMA-2-70B-Chat~\cite{touvron2023llama2} on \dataname without providing the documents.
In this way, the model can only rely on its memory to answer the questions.
As figure~\ref{fig:llama-2-70b-chat-wo-psg} shows, the accuracy of LLaMA-2-70B-Chat increases as the topic entity becoming more salient, which testifies that the saliency of entities in Wikidata5M is highly correlated with the memory of mainstream LLMs.

\section{Experiment Setup}

In this section, we introduce the evaluated model and the evaluated knowledge conflict resolving methods.
We also setup the evaluation metric.

\subsection{Implemented Models}

We split evaluated LLMs into two categories: (1) \textbf{Pre-trained Only LLMs}, including encoder-decoder models---UL2 (20B) \cite{tay2023ul2}---and decoder-only models---GPT-J (6B) \cite{gpt-j}, GPT-NeoX (20B) \cite{black2022neox}, LLaMA (7B, 13B, 30B, 65B) \cite{touvron2023llama} and LLaMA-2 (7B, 13B, 70B) \cite{touvron2023llama2}.
(2) \textbf{Assistant LLMs}, which are further fine-tuned to align to human instructions, including Flan-T5 (11B), Flan-UL2 (20B) \cite{2022flant5}, GPT-JT (6B)~\cite{gpt-jt}, GPT-NeoXT-Chat-Base (20B) \cite{to2023neoxchat}, Alpaca (7B,13B) \cite{taori2023alpaca}, Vicuna (7B,13B) \cite{vicuna2023}, Tulu (7B,13B,30B) \cite{wang2023tulu}, LLaMA-2-Chat (7B, 13B, 70B) \cite{touvron2023llama2}, and 
variants of InstructGPT \cite{ouyang2022training}.
They are denoted as \texttt{text-curie-001} (6.7B), \texttt{text-davinci-002} (175B), \texttt{gpt-3.5-turbo}
\footnote{The model size of \texttt{gpt-3.5-turbo} is unclear.},
 and \texttt{text-davinci-003} (175B).

We evaluate the overall performance of these LLMs in the setting of one-shot prompting. 
Specially, we prepare 10 human-annotated examples for each level of \dataname. 
To mitigate the sensitivity of example selection in in-context learning of LLMs, we randomly choose one example for each data in \dataname.
As one-shot prompting prevents LLMs from learning knowledge conflict resolution skills from more than $1$ training data, the achieved performance approximates the intrinsic capacity to resolve knowledge conflicts.

\subsection{Evaluation Metric} 
\label{sec:metric}

Considering the flexibility of the output of LLMs, The prediction is marked correct when it matches any of the aliases of the answer entity from Wikidata.
We use accuracy as the evaluation metric.

In particular, to validate the feasibility of this metric, we randomly sample 100 questions from \dataname and corresponding answers from 22 different LLMs. 
Then we ask human annotators to label the correctness of the answers. 
We calculate the correlation between human-evaluation score and the proposed alias-based evaluation accuracy for each LLMs with regard to the sampled questions.
The correlation coefficient is $0.87$, which indicates that the automatic score is a good estimation for model performance in our task.

\section{Results and Analysis}

\subsection{Overall Results}
\label{sec:overall_results}

We report one-shot prompting accuracy in Table~\ref{tab:main-result} as the overall results, which keeps the minimum volume of exposed training data to the LLMs.
Overall, we have $3$ observations:
(1) Mainstream LLMs, especially LLaMA-2-Chat and InstructGPT variants, achieve satisfactory results on \datanameS, which indicates that LLMs have the capacity to extract conflicting knowledge directly.
(2) However, performance drops by a large margin when reasoning with conflicting knowledge in \datanameE and \datanameI is required, which suggests that advanced utilization of conflicting knowledge is still challenging for LLMs. 
(3) Compared to pre-trained only LLMs, \textit{i.e.,} LLaMA, LLaMA-2, GPT-J, GPT-NeoX, UL2 and T5, their counterpart assistant models \textit{i.e.,} Alpaca, LLaMA-2-Chat, GPT-JT, GPT-NeoXT, Flan-UL2 and Flan-T5, promise an improvement in performance, due to their better understanding of the instructions, which requires utilizing the supplemented document.

Intriguingly, we find that the 13B variant of LLaMA and LLaMA-2 performs worse than the 7B variant.
By further examining the output from 13B variants, we find that they prefer continuing asking similar questions to answering the question.
This observation indicates that scaling up pre-trained only LLMs can enhance the ability to continue writing, but the level of knowledge mastery is not guaranteed to improve accordingly, nor are the task-solving and instruction-following abilities. 
By contrast, this is not observed in Alpaca-13B, Vicuna-13B,Tulu-13B and LLaMA-2-13B-Chat which suggests that instruction alignment tuning helps to resolve knowledge conflicts by improving the ability to follow instructions.

\begin{table}[t]
    \centering
    \setlength{\tabcolsep}{3pt}
    \scalebox{0.86}{
    \begin{tabular}{llrrrr}
    \toprule
    Type & LLM      & \datanameS & \datanameE & \datanameI \\ \midrule
    \multirow{10}{*}{\rotatebox{90}{Pre-trained\;\ Only}}
    & UL2      & $52.66$  & $13.03$   & $3.33$   \\
    & GPT-J    & $48.57$  & $29.84$   & $18.63$   \\
    & GPT-NeoX   & $55.65$  & $30.72$   & $20.39$   \\
    & LLaMA-7B    & $64.19$  & $38.20$   & $21.18$   \\
    & LLaMA-13B    & $49.65$  & $27.20$   & $17.45$   \\
    & LLaMA-30B    & $70.36$  & $40.49$   & $23.33$   \\
    & LLaMA-65B    & $67.89$  & $38.82$   & $24.51$   \\
    & LLaMA-2-7B   & $63.88$  & $35.65$   & $20.59$   \\
    & LLaMA-2-13B  & $77.18$  & $42.52$   & $24.51$   \\
    & LLaMA-2-70B  & $59.63$  & $32.75$   & $20.20$   \\
    \cmidrule(lr){1-5}
    \multirow{17}{*}{\rotatebox{90}{Assistant\;\ LLMs}}
    & Flan-T5-XXL    & $85.95$  & $40.67$   & $15.69$   \\
    & Flan-UL2    & $84.23$  & $46.48$   & $15.29$   \\
    & GPT-JT   & $60.59$  & $32.57$   & $17.25$   \\
    & GPT-NeoXT   & $70.03$  & $38.73$   & $27.65$   \\
    & Alpaca-7B    & $82.74$  & $46.83$   & $29.22$   \\
    & Alpaca-13B    & $85.82$  & $43.75$   & $28.82$   \\
    & Vicuna-7B    & $75.92$  & $42.43$   & $26.67$   \\
    & Vicuna-13B    & $84.20$   & $46.30$    & $25.29$   \\
    & Tulu-7B     & $86.72$  & $51.50$    & $27.06$   \\
    & Tulu-13B    & $87.34$  & $53.26$   & $26.67$   \\
    & Tulu-30B    & $89.61$  & $53.61$   & $24.31$   \\
    & LLaMA-2-7B-Chat   & $90.66$  & $50.18$   & $28.82$   \\
    & LLaMA-2-13B-Chat  & $91.66$  & $53.43$   & $35.88$   \\
    & LLaMA-2-70B-Chat  & $94.93$  & $58.54$   & $38.24$   \\
    \cmidrule(lr){2-5}
    & \texttt{text-curie-001}   & $74.22$  & $42.96$   & $29.41$   \\
    & \texttt{text-davinci-002}  & $87.45$  & $55.02$   & $30.59$   \\
    & \texttt{gpt-3.5-turbo}   & $83.66$  & $34.95$   & $10.20$   \\
    & \texttt{text-davinci-003}  & $95.11$  & $63.56$   & $35.49$   \\
    \bottomrule
    \end{tabular}
    }
    \caption{One-shot prompting results on \dataname. All the results are presented in accuracy (\%).}
    \label{tab:main-result}
\end{table}

Moreover, although Flan-T5 and Flan-UL2 present competitive performance on \datanameS, they fall short of other assistant language models on \datanameI.
This is because Flan-T5 and Flan-UL2 are fine-tuned with existing datasets such as CommonsenseQA~\cite{talmor-etal-2019-commonsenseqa}, while the instruction datasets of Alpaca and Vicuna are generated from analogous or real dialog between human and LLMs, which is more diverse.
Therefore, it is necessary to diversify instruction datasets. 

\begin{table}[t]
   \scalebox{0.8}{
   \renewcommand{\arraystretch}{1.2}
   \setlength{\tabcolsep}{3pt}
   \begin{tabular}{lrrrrrr}
   \toprule
    & LLMs                         & \datanameS & \datanameE & \datanameI \\ 
   \midrule
   
   \multirow{4}{*}{\rotatebox{90}{Instr+Opin}}
   
   & LLaMA-7B            & \cellcolor{'sred'} $57.0$ ($-\;\;7.2$)   & \cellcolor{'sgreen'} $40.9$ ($+\;\;2.7$)   & \cellcolor{'sred'}$17.7$ ($-\;\;3.5$)    \\
   & Alpaca-7B           & \cellcolor{'sred'} $78.2$ ($-\;\;4.6$)   &  $46.8$ ($\;\;\;\;0.0$)                        & \cellcolor{'sred'} $26.1$ ($-\;\;3.1$)    \\
   & Vicuna-7B           & \cellcolor{'sred'} $32.6$ ($-43.4$)  & \cellcolor{'sred'} $25.0$ ($-17.4$)    & \cellcolor{'sred'} $12.9$ ($-13.7$)       \\
   & Tulu-7B             & \cellcolor{'sred'} $83.5$ ($-\;\;3.3$)   & \cellcolor{'sred'} $47.6$ ($-\;\;3.9$)     & \cellcolor{'sred'} $26.3$ ($-\;\;0.8$)   \\ 
     &Tulu-13B             & \cellcolor{'sred'} $88.8$ ($+\;\;1.5$)   & \cellcolor{'sred'} $46.8$ ($-\;\;6.5$)     & $26.7$ ($\;\;\;\;0.0$)   \\
   \midrule
   \multirow{4}{*}{\rotatebox{90}{Instr+Attr}}
   & LLaMA-7B             & \cellcolor{'sred'} $63.0$ ($-\;\;1.2$)  & \cellcolor{'sgreen'} $41.7$ ($+\;\;3.5$)    & \cellcolor{'sred'} $20.8$ ($-\;\;0.4$)    \\
   & Alpaca-7B            & \cellcolor{'sred'} $81.4$ ($-\;\;1.4$)  & \cellcolor{'sgreen'} $48.7$ ($+\;\;1.9$)    & \cellcolor{'sred'} $25.7$ ($-\;\;3.5$)    \\
   & Vicuna-7B            & \cellcolor{'sred'} $43.2$ ($-32.8$) & \cellcolor{'sred'} $34.5$ ($-\;\;7.9$)      & \cellcolor{'sred'} $12.4$ ($-14.3$)       \\
   & Tulu-7B              & \cellcolor{'sred'} $82.6$ ($-\;\;4.1$)  & \cellcolor{'sred'} $49.6$ ($-\;\;1.9$)      & \cellcolor{'sred'} $24.1$ ($-\;\;2.9$)   \\ 
     & Tulu-13B             & \cellcolor{'sred'} $88.2$ ($+\;\;0.9$)  & \cellcolor{'sred'} $46.9$ ($-\;\;6.3$)      & \cellcolor{'sred'} $24.7$ ($-\;\;2.0$)   \\
   \bottomrule
   \end{tabular}
   }
   \caption{
   Accuracy (\%) with prompting strategies.
   We show absolute accuracy changes compared with results from Table~\ref{tab:main-result} in parenthesis.
   }
   \label{tab:prompting}
   \end{table}

\subsection{Conflicting Knowledge Resolving Methods Analysis}

We thoroughly explore $3$ main existing conflict-resolving aspects, \textbf{Prompting}, \textbf{Decoding} and \textbf{Fine-tuning}.

\textbf{Prompting} is the most common way to fit LLMs into down-stream tasks as it is straightforward and easy to use. 
In terms of the conflicting knowledge resolving task, \citeauthor{chen2023context} design two elaborated prompting strategies to improve the ability of LLMs to handle knowledge conflicting problem.

\textbf{Decoding} is a more subtle way to control the output of LLMs compared to prompting. 
Some previous work~\cite{welleck2019neural,welleck2020consistency} has proved text generated with proper decoding strategies like beam-search to be more fluent. 
For resolving conflicting knowledge, \citet{shi2023cad} recently propose \textbf{Context Aware Decoding (CAD)} by calibrating the output logits, which forces LLMs to pay more attention to the provided document.

\textbf{Fine-Tuning} is the most direct way to change the behaviour of LLMs.
Recent studies~\cite{zhou2023lima} point out that a small-scale well-curated supervised training dataset is enough to teach models to produce high quality output. 
Following this idea, we fine-tune LLMs on the training set.

\subsubsection{Prompting Analysis}

We use regular question answering prompting (\textit{read the passage and answer the following question}) as the baseline methods.
We report the performance of two prompting strategies proposed by \citet{chen2023context} (\textit{i.e.,} Instr+Opin and Instr+Attr) as well as their absolute performance improvement \textit{w.r.t.} baseline prompt in Table~\ref{tab:prompting}.

The results do not present a promising improvement in performance. 
We argue that the reason is that LLMs are sensitive to the prompting strategy.
One well-designed strategy on one dataset may not be suitable for another dataset.
The huge performance drop of Vicuna-7B indicates that for the dialogue assistant LLMs, which are fine-tuned with specific prompt templates, have difficulties in adopting our prompt strategy directly.

\subsubsection{Decoding Strategy Analysis}

We use greedy search as the baseline decoding strategy to examine the effectiveness of context-aware decoding~\cite{shi2023cad}.
The performance of CAD and its absolute improvement is shown in Table~\ref{tab:cad}.

The results show that CAD helps LLMs to trust the supplement document and achieve better performance in \datanameS.
However, we hypothesize that \textit{CAD diminishes LLMs' ability to access intrinsic knowledge}.
When the question requires both the conflicting knowledge, and related information in the memory, as required by \datanameE and \datanameI, performance drops inevitably.
The experiment results testifies our hypothesis.

\begin{table}[t]
\renewcommand{\arraystretch}{1.2}
\scalebox{0.9}{
\setlength{\tabcolsep}{3pt}
\begin{tabular}{lrrrrr}
\toprule
LLMs                         & \datanameS & \datanameE & \datanameI \\ 
\midrule
\multirow{1}{*}{LLaMA-7B}              
   & \cellcolor{'sgreen'} $66.5$ ($+2.3$)  & \cellcolor{'sred'} $37.1$ ($-1.1$)    & \cellcolor{'sred'}$9.2$ ($-12.0$)    \\
\multirow{1}{*}{Alpaca-7B}           
   &  $82.7$ ($\;\;\;0.0$)  & \cellcolor{'sgreen'} $50.3$ ($+3.4$)    & \cellcolor{'sred'} $20.4$ ($-\;\;8.8$)    \\
\multirow{1}{*}{Vicuna-7B}            
   & \cellcolor{'sgreen'} $77.3$ ($+1.4$)   & \cellcolor{'sred'} $41.6$ ($-0.8$)    & \cellcolor{'sred'} $18.8$ ($-\;\;7.9$)       \\
\multirow{1}{*}{Tulu-7B}             
   & \cellcolor{'sgreen'} $87.7$ ($+0.9$)  & \cellcolor{'sred'} $42.9$ ($-8.6$)     & \cellcolor{'sred'} $11.4$ ($-15.7$)   \\ 

\multirow{1}{*}{Tulu-13B}
   & \cellcolor{'sgreen'} $91.5$ ($+4.2$)  & \cellcolor{'sred'} $44.3$ ($-9.0$)     & \cellcolor{'sred'} $15.9$ ($-10.8$)   \\
                           \bottomrule
\end{tabular}
}
\caption{
Accuracy (\%) with context-aware decoding.
We show absolute accuracy changes compared with results from Table~\ref{tab:main-result} in parenthesis.
}
\label{tab:cad}
\end{table}

\subsubsection{Fine-Tuning Analysis}

Previous analysis show that there is no silver bullet that resolves knowledge conflicts with LLMs without fine-tuning.
In order to further address the issue of knowledge conflicts that LLMs may encounter, we employ human annotated data to fine-tune LLMs to resolve knowledge conflicts.

\textbf{Fine-Tuning Setup.}
The general purpose of fine-tuning LLMs with the annotated data is two-fold.
(1) To enable LLMs to understand instructions to resolve knowledge conflicts;
(2) To enable LLMs to generate rationale as the chain-of-thought in more complicated reasoning.

To prevent catastrophic forgetting, we mix the fine-tuning dataset with Wikitext~\cite{merity2016wikitext} and other supervised fine-tuning dataset, \textit{e.g.,} Alpaca dataset~\cite{taori2023alpaca}.
Finally, we fine-tune the 7B version of LLaMA, Alpaca, Vicuna, and Tulu on the mixed data.
We train these models for two epochs at a learning rate of $2\times10^{-5}$.
We use 3\% of the total training steps for linear learning rate warmup and adopt a linear learning rate decay.

\begin{table}[t]
\renewcommand{\arraystretch}{1.2}
\centering
\setlength{\tabcolsep}{3pt}
\scalebox{0.88}{
\begin{tabular}{lrrrr}
\toprule
LLMs                & \datanameS & \datanameE & \datanameI \\ 
\midrule
    \multirow{1}{*}{LLaMA-7B}              
       & \cellcolor{'sgreen'} $90.1$ ($+25.9$)  & \cellcolor{'sgreen'} $60.2$ ($+22.0$)    & \cellcolor{'sgreen'} $32.0$ ($+10.8$)    \\
    \multirow{1}{*}{Alpaca-7B}           
       & \cellcolor{'sgreen'} $92.5$ ($+\;\;9.8$)  & \cellcolor{'sgreen'} $59.2$ ($+12.4$)    & \cellcolor{'sgreen'} $32.2$ ($+\;\;2.9$)    \\
    \multirow{1}{*}{Vicuna-7B}            
       & \cellcolor{'sgreen'} $94.6$ ($+18.6$)   & \cellcolor{'sgreen'} $60.0$ ($+17.6$)    & \cellcolor{'sgreen'} $32.9$ ($+\;\;6.3$)       \\
    \multirow{1}{*}{Tulu-7B}             
       & \cellcolor{'sgreen'} $90.2$ ($+\;\;3.4$)  & \cellcolor{'sgreen'} $54.1$ ($+\;\;2.6$)     & \cellcolor{'sgreen'} $27.5$ ($+\;\;0.4$)   \\ 
   \multirow{1}{*}{Tulu-13B}
      & \cellcolor{'sgreen'} $93.5$ ($+\;\;6.2$)  & \cellcolor{'sgreen'} $56.7$ ($+\;\;3.4$)     & \cellcolor{'sgreen'} $31.2$ ($+\;\;4.5$)   \\
\bottomrule
\end{tabular}
}
\caption{
Accuracy (\%) and their absolute changes (compared with Table~\ref{tab:main-result}) after fine-tuning.
}
\label{tab:train}
\end{table}

\textbf{Fine-Tuning Results.}
The experiment results after fine-tuning LLMs are shown in Table~\ref{tab:train}.
Generally, a promising performance improvement from the extra training process can be observed.
LLaMA-7B has more performance gain compared with the assistant LLMs.
It even reaches a comparable capability to resolve knowledge conflicts with complicated reasonings.
Moreover, after fine-tuning, Vicuna-7B reaches a comparable result with LLaMA-2-70B-Chat in Table~\ref{tab:main-result}, indicating the role of fine-tuning relatively small scale models in resolving knowledge conflicts as the fallback.

\subsection{Effect of Reasoning Types}
\label{sec:reasoning}

We aim to understand how LLMs react differently to conflicting knowledge across 3 tasks: direct extraction in \datanameS, explicit reasoning in \datanameE, and implicit reasoning in \datanameI. 
To this end, we evaluate the tendency of LLMs to rely on their inherent parametric knowledge $k^\prime$ versus adapting to the conflicting knowledge $k$ provided in the context.
Particularly, we denote the set of aliases of answer to the question when consider parametric knowledge $k^\prime$, as $\{a_i^\prime\}$.
Following the same metric we introduce in Section~\ref{sec:metric},
we evaluate the accuracy of LLM with regard to $\{a_i'\}$.
This accuracy score thus reflects LLMs' tendency for shortcut reasoning based on their parametric knowledge, 
rather than carefully reading the conflicting knowledge to deduce the answer accordingly.

Our analysis includes both pre-trained only LLMs (LLaMA-2-7B, LLaMA-2-13B and LLaMA-2-70B) and assistant LLMs that are further developed with Reinforcement Learning from Human Feedback (RLHF) (LLaMA-2-7B-chat, LLaMA-2-13B-chat and LLaMA-2-70B-chat) under one-shot setting, with results presented in Tables~\ref{tab:origin-answer}.
There are three main observations:

\textbf{Knowledge conflicting with direct extraction is almost resolved by mainstream LLMs.} 
As in Table~\ref{tab:origin-answer}, LLaMA-2-13B displays the highest score, $1.22$, in answering the question directly based on parametric knowledge on \datanameS.
Meanwhile, the accuracy of all the RLHF variants of the LLaMA-2 surpass $90\%$, as shown in Table~\ref{tab:main-result}.
These results imply that LLMs, especially assistant LLMs, are able to resolve knowledge conflicts with direct extraction.
This is because direct extraction only requires LLMs to extract the answer from the context, which is a common align task in developing assistant LLMs.
Therefore, LLMs are able to extract the conflicting knowledge $k$ from the context and ignore the parametric knowledge $k^\prime$.

\begin{table}[t]
    \centering
    \scalebox{0.93}{
    \setlength{\tabcolsep}{4pt}
    \begin{tabular}{lccc}
    \toprule
    {LLMs} & {\datanameS} & {\datanameE} & {\datanameI} \\
    \midrule
    LLaMA-2-7B        & $0.76$ & $14.79$ & $15.69$ \\
    LLaMA-2-13B       & $1.22$ & $14.00$ & $15.88$ \\
    LLaMA-2-70B       & $0.65$ & $13.83$ & $14.90$ \\
    LLaMA-2-7B-chat   & $0.42$ & $14.88$ & $18.24$ \\
    LLaMA-2-13B-chat  & $0.53$ & $13.47$ & $23.33$ \\
    LLaMA-2-70B-chat  & $0.57$ & $13.38$ & $28.24$ \\
    \bottomrule
    \end{tabular}
    }
    \caption{The Accuracy (\%) of LLMs w.r.t $\{a_i^\prime\}$ on \dataname.
    $\{a_i^\prime\}$ is the set of aliases of answer to the question when the parametric knowledge $k'$ is considered, rather than the conflicting knowledge $k$.
    }
    \label{tab:origin-answer}
\end{table}

\textbf{Resolving conflicting knowledge with complex reasoning requirement stimulates LLMs to stick to their parametric knowledge.}
We hypothesize that LLMs, instead of performing complex reasoning, tend to find a shortcut to answer the question, which is to sticking to their parametric knowledge $k^\prime$.
It is obvious from Table~\ref{tab:main-result} that the accuracy of LLMs on \datanameE and \datanameI is significantly lower than that on \datanameS. 
Since the answer of \datanameE and \datanameI does not appear in the context, LLMs are required to perform complex reasoning to find conflicting knowledge $k$ as clues to answer the question.
Meanwhile, both \datanameE and \datanameI have a higher chance than \datanameS in generating $e_a'$ as the answer, as shown in Table~\ref{tab:origin-answer}.
These phenomena validate our hypothesis.

\textbf{The impact of scaling up LLMs is double-edged: While the stickiness to parametric knowledge mildly alleviates in \datanameE, it exacerbates in \datanameI.}
As shown in Table~\ref{tab:origin-answer}, the accuracy of LLMs w.r.t the answer $\{a_i'\}$ from parametric knowledge $k'$ decreases as the size of LLMs increases for \datanameE, but increases for \datanameI.
This phenomenon mainly roots in the difference between explicit and implicit reasoning skills.
Although \datanameE and \datanameI share the feature that answers cannot be directly extracted from their context, questions in \datanameE provide explicit reasoning paths, while \datanameI calls for LLMs to seek out the reasoning path by themselves.
As size escalates in LLMs, they are more capable of following reasoning path provided in questions.
Therefore, the accuracy w.r.t $\{a_i'\}$ decreases as the size of LLMs increases for \datanameE.
Meanwhile, as LLMs increase in size, they accumulate parametric knowledge, which fosters their inclination to find shortcuts to answer in \datanameI, instead of seeking the reasoning path by themselves, which shows more complexity.

\section{Related Works}
There are mainly two threads of works focusing on conflicting knowledge. 
One thread identifies various tasks associated with conflicting knowledge, while the other thread concentrates on context-faithful generation with the presence of conflicting knowledge.

\textbf{Conflicting Knowledge Tasks.} 
Research into knowledge conflicts can be traced back to studies on backdoor attacks, where LLMs are susceptible to errors when provided conflicting knowledge~\cite{niu2018adversarial,kassner2021beliefbank,du2022synthetic}.
More recently, knowledge conflicts, which has been demonstrated by entity conflicting~\cite{longpre2021entity} or relation conflicting~\cite{chen2023context} in the form of question answering~\cite{pan2021contraqa}, stands in the spotlight.
This also occurred in downstream applications, such as knowledge-grounded dialogue systems~\cite{glmdialogue}.
However, these works often neglect the necessity of complex reasoning processes with conflicting knowledge.

\textbf{Context-faithful Generation} is a number of strategies adopted to resolve conflicting knowledge in generative LLMs.
One strategy is constraint decoding, where LLMs generate answers from a constrained dictionary~\cite{post2018fast,welleck2019non,nye2021improving}.
Prompting-based methods design sophisticated prompt templates to persuade LLMs to adhere faithfully to the provided document~\cite{si2022prompting,chen2023context}.
Decoding-based methods modify the output logits of each generated token conditioned on the provided documents~\cite{shi2023cad,chen2022rich}.
However, context-faithful generation methods only take effect when advanced reasoning is not required.

\section{Conclusion and Future Directions}

This paper introduces \dataname, a dataset designed to examine conflicting knowledge resolution capability when advanced reasoning skills are required. 
The dataset does not just assess the ability to extract counter-parameter knowledge, but also evaluates the capacity to reason with conflicting knowledge. 
We conduct evaluation on mainstream LLMs alongside existing methods for resolving knowledge conflicts, culminating in guidelines for selecting an appropriate solution for knowledge conflict resolution.

In the future, we will explore a broader range of knowledge conflicting scenarios, such as counter-commonsense knowledge.
Apart from fine-tuning, we are also interested in developing more lightweight solutions to resolve knowledge conflicts in a more general condition.

\section*{Code and Data Availability Statement}

The artifacts associated with this paper include both datasets and experiment codes.

For dataset construction, it is desirable to enable humans to generate the documents and to generate the questions.
However, this approach is extremely time-consuming and labor-intensive. 
To strike a balance between cost and quality, we employ human annotators to label the training data and filter out low-quality data points for the test set.
As we demonstrate in Section~\ref{sec:metric}, evaluation results on the automatically generated data are consistent with human evaluation, satisfying our requirements.
Thus, the whole dataset include (1) automatically generated dataset; (2) human annotated dataset; and (3) human filtered dataset.

Our codes include scripts that are used to generate data, and codebases that are used to establish the initial baselines.

Per the request of anonymous protocol, we will release both the automatically generated dataset, human annotated dataset, and human filter dataset as soon as this paper is accepted after the reviewing process.

\section*{Ethical Consideration}
We primarily focus on two ethical issues: the treatment of annotators and the usage of data resources.

For the treatment of annotators, they are all students undertaking post-secondary educations.
The annotator work is conducted in their part-time.
We provide hourly wages that meet the local average for data work, ensuring equal pay for equal work across all employees.

Data resources include both the checkpoint of models and the training data. 
We use these data in accordance with the requirements of the data providers, and this research has no commercial intentions. 
The usage of this portion of data is purely for academic purposes.

\section*{Limitations}
We would like to highlight that our data annotation process relies on the help of \texttt{gpt-3.5-turbo}. Although it has been shown to be effective in the synthesis of datasets by appropriately prompting large language models~\cite{taori2023alpaca,vicuna2023}, there are also observations pointing out that this could potentially introduce bias.
To resolve this bias, we hire and train a group of annotators with secondary education experience to filter out low-quality data and construct $850$ dataset from scratch (which means they are generated by human rather than LLMs).

\section*{Acknowledgement}
This work is supported by a grant from the Institute for Guo Qiang, Tsinghua University (2019GQB0003), Tsinghua University Initiative Scientific Research Program, and Zhipu AI.

\section*{Bibliographical References}
\bibliography{1-ref}
\bibliographystyle{lrec_natbib}

\clearpage

\appendix

\section{Prompts for Data Construction}

\subsection{Document Generation}
\label{app:document_gen}

We use the following prompt to generate document from the ego network.

\begin{table}[h]
    \centering
    \scalebox{0.72}{
    \begin{tabular}{p{1.3\linewidth}}
        \toprule

            \vspace{-2mm}
        \textbf{\textsc{Instruction:}} Image you are a journalist. And you are given some semantic triples from Wikidata about one specific topic. And you are planning to write one short passage provided to your reader based on the information that these semantic triples. As its name indicates, a semantic triple is a set of three entities that codifies a statement about semantic data in the form of subject–predicate–object expressions (e.g., "Bob is 35", or "Bob knows John"). Your task is to convert these semantic triples into a concise and coherent passage with approximately 200 words to help your readers understand the specific topic.

Attention 1: In the resulting passage, DO NOT USE any text directly contains entity, semantic triple, or Wikidata, etc.

Attention 2: In the resulting passage, all the statements presented in the given semantic triples should be included.

Attention 3: If there is any counterfactual info contained in the given semantic triples, you should just accept it and faithfully describe this counterfactual info. DO NOT USE any words like "interesting", or "surprising" to indicate this info is counterfactual to passage readers. \\
        \midrule
               \textbf{\textsc{Input:}} \\
Semantic Triples From Wikidata:\\
(Lionel Messi, place of birth, Rosario) \\
(Lionel Messi, sex or gender, male) \\
(Lionel Messi, date of birth, 24 June 1987) \\
(Lionel Messi, member of sports team, Inter Miami CF) \\
(Lionel Messi, occupation, Footballer) \\
(Lionel Messi, position played on team / speciality, Forward) \\
(Lionel Messi, position played on team / speciality, Midfielder) \\
(Lionel Messi, award received, Ballon d'Or) \\
(Lionel Messi, award received, FIFA World Player of the Year) \\
(lionel Messi, sponsor, Adidas)\\
        \midrule
               \textbf{\textsc{Output:}} Lionel Messi, hailing from the city of Rosario, was born on the 24th of June, 1987. This seasoned footballer, known for his exceptional skills and tenacity on the field, is now a key player for Inter Miami CF. Although predominantly recognized as a forward, Messi has also showcased his versatility by frequently occupying the midfielder position. The Argentine star's performance over the years has not gone unnoticed. His trophy cabinet includes accolades such as the Ballon d'Or and the FIFA World Player of the Year, testifying to his profound impact on the sport. Besides his sports-related commitments, Messi has also embarked on commercial ventures, endorsing global brands like Adidas. His enduring legacy continues to influence and reshape the world of football.\\
        \bottomrule
    \end{tabular}
    }
        \caption{
  The instruction and an example of Document Generation
    }
    \label{tab:docgen}
\end{table}

\newpage
\subsection{Question Generation}

\subsubsection{Question Generation for \datanameS and \datanameI}
As questions in both \datanameS and \datanameI are equivalent to a triple query, we use the following prompt to generate these questions.
\begin{table}[h]
    \centering
    \scalebox{0.72}{
    \begin{tabular}{p{1.3\linewidth}}
        \toprule

            \vspace{-2mm}
        \textbf{\textsc{Instruction:}} A semantic triple describe the relation between one head entity and one tail entity. For example, Job Biden -> native language -> English is one semantic triple which means Job Biden (head entity)’s native language (relation) is English (tail entity), now you are given one incomplete semantic triple where the tail entity is missing and one hint which would tell what all the possible missing entity is. your task is to design 5 questions based on the given semantic triple and the hint to find out the missing tail entity. 
Notice: the given hint could be utilized to design more accurate questions with respect to the given possible missing entity, but any part of the hint should not be contained in the generated question!
\\
Example:\\
Input:
incomplete semantic triple: Faiyum -> 	
located in the administrative territorial entity (the item is located on the territory of the following administrative entity.) -> missing entity
Hint: the missing entity is "Egypt"

Output: \\
1. Which country is Faiyum a part of? \\
2. What is the sovereign state that contains Faiyum? \\
3. Faiyum is in which country? \\
4. In which country would I be if I was visiting Faiyum? \\
5. If I were to send a letter to Faiyum, which country's name should I write on the envelope? \\

Example 2:
\\
        \midrule
               \textbf{\textsc{Input:}} 	incomplete semantic triple:Paradise ->Country (sovereign state that this item is in (not to be used for human beings)) -> missing entity\\
        \midrule
               \textbf{\textsc{Output:}}\\ 1. Which sovereign state does Paradise belong to?\\
2. Is Paradise located in a province or a territory of Canada?\\
3. What is the name of the country where Paradise is located?\\
4. Which country has Paradise as a part of its sovereign state?\\
5. If I were to travel to Paradise, which country would I need to visit\\
        \bottomrule
    \end{tabular}
    }
        \caption{
  The instruction and an example of One hop question generation.
    }
    \label{tab:qg1}
\end{table}

\clearpage
\subsubsection{Multi-hop question generation for \datanameE}

We use the following prompt to generate questions explicitly stating multiple hops in \datanameE.

\begin{table}[h]
    \centering
    \scalebox{0.72}{
    \begin{tabular}{p{1.3\linewidth}}
        \toprule

            \vspace{-2mm}
        \textbf{\textsc{Instruction:}} You are given one multi-hop reasoning chain over Wikidata. This chain is related to a multi-hop KGQA question. Your task is to generate several multi-hop KGQA questions in natural language based on the given multi-hop reasoning chain. \\

Input 1: Michel Verne -> place of birth (P19) -> Middle Entity -> country (P17) -> Answer Entity \\
Output 1: \\
    1. Which country does Michel Verne's birth city belong to? \\
    2. In which country can we find the city where Michel Verne was born \\
    3. To which country is the city of Michel Verne's birthplace associated with? \\
    4. The city where Michel Verne was born is in which country? \\
    5. What is the nationality of the country where Michel Verne was born? \\

Input 2: Buick LaCrosse -> brand (P1716) -> Middle Entity -> location of formation (P740) -> Answer Entity\\
Output 2: \\
    1. In which city was the company that created the Buick LaCrosse brand established?\\
    2. What is the city of origin of the brand associated with the Buick LaCrosse?\\
    3. The brand that owns Buick LaCrosse was formed in which city?\\
    4. Where was the company that produced the Buick LaCrosse initially established?\\
    5. The Buick LaCrosse is from a brand that was formed in what city? \\
        \midrule
               \textbf{\textsc{Input 3:}} 	Lionel Messi -> Member of sports team (P54) -> Middle Entity -> League (P118) -> Answer Entity\\
        \midrule
               \textbf{\textsc{Output 3:}} \\ 1. What league is Lionel Messi's sports team a part of? \\
2. In which league does Lionel Messi's sports team compete? \\
3. To which league does the sports team Lionel Messi belongs to belong? \\
4. Which league is associated with Lionel Messi's sports team? \\
5. What is the name of the league that Lionel Messi's sports team plays in?\\
        \bottomrule
    \end{tabular}
    }
        \caption{
  The instruction and an example of Two hop question generation.
    }
    \label{tab:qg2}
\end{table}

\newpage
\section{Prompts for Model Evaluation}

\subsection{Prompt Template for InstructGPT variants, LLaMA, GPT-J, GPT-JT, and GPT-NeoX}

For pre-trained only models, they are not aligned to a specific format of instructions.
We use a unified prompting template for InstructGPT variants, LLaMA, GPT-J, GPT-JT, and GPT-NeoX.

\begin{table}[h]
    \centering
    \small
    \begin{tabular}{p{0.85\linewidth}}
        \toprule

            \vspace{-2mm}
        \textbf{\textsc{Instruction:}} \\ read the passage and answer the question. You may do some analysis on the passage and question before answering. But you should directly give out in the answer in the end and in the form of {{answer}}. \\
        \midrule
               \textbf{\textsc{Examples:}}\\
        Passage: [[example\_passage\_1]]\\
Question: [[example\_question\_1]]\\
Answer: [[example\_answer\_1]]\\
        ... \\
        \midrule
               \textbf{\textsc{Example K+1:}} 
Passage: [[passage]]\\
Question: [[question]]\\
Answer: \\
        \bottomrule
    \end{tabular}
        \caption{
  Prompt template for InstructGPT variants, LLaMA, GPT-J, GPT-JT, and GPT-NeoX.
    }
   \end{table}

\newpage
\subsection{Prompt Template for Alpaca}

We use the format in the supervised fine-tuning data of Alpaca~\cite{taori2023alpaca}.

\begin{table}[h]
    \centering
    \small
    \begin{tabular}{p{0.85\linewidth}}
        \toprule

            \vspace{-2mm}
        \textbf{\textsc{Instruction:}} \\
        Below is an instruction that describes a task, paired with an input that provides further context. Write a response that appropriately completes the request.

\#\#\# Instruction:
Instruction: Read the passage and answer the question. You may do some analysis on the passage and question before answering. But you should directly give out in the answer in the end and in the form of {{answer}}. \\
        \midrule
               \textbf{\textsc{Examples:}}\\
        \#\#\# Input 1:
Passage: [[example\_passage\_1]] \\ 
Question: [[example\_question\_1]] \\

\#\#\# Response 1:
Answer: [[example\_answer\_1]] \\
        ... \\
        \#\#\# Input K: \\ 
Passage: [[example\_passage\_K]] \\
Question: [[example\_question\_K]] \\

\#\#\# Response K: \\
Answer: [[example\_answer\_K]] \\
        \midrule
               \textbf{\textsc{Example K+1:}}  \\
\#\#\# Input K+1
Passage: [[passage]]\\
Question: [[question]]\\
\#\#\# Response K+1: \\
        \bottomrule
    \end{tabular}
        \caption{
  Prompt template for Alpaca.
    }
   \end{table}

\newpage
\subsection{Prompt Template for Vicuna}

We use the format in the supervised fine-tuning data of Vicuna~\cite{vicuna2023}.

\begin{table}[h]
    \centering
    \small
    \begin{tabular}{p{0.85\linewidth}}
        \toprule

            \vspace{-2mm}
        \textbf{\textsc{Instruction:}} \\ A chat between a curious user and an artificial intelligence assistant. The assistant gives helpful, detailed, and polite answers to the user's questions. USER: Read the passage and answer the question. You may do some analysis on the passage and question before answering. But you should directly give out in the answer in the end and in the form of {{answer}}. ASSISTANT: OK, I got it.</s> \\
        \midrule
               \textbf{\textsc{Examples:}}\\
        USER: Passage: [[example\_passage\_1]] Question: [[example\_question\_1]] ASSISTANT: Answer: [[example\_answer\_1]]</s>\\
        ...\\
        USER: Passage: [[example\_passage\_K]] Question: [[example\_question\_K]] ASSISTANT: Answer: [[example\_answer\_K]]</s>\\ 
        \midrule
        \textbf{\textsc{Example K+1:}}  \\
        USER: Passage: [[passage]] Question: [[question]] ASSISTANT: Answer: \\
        \bottomrule
    \end{tabular}
        \caption{
  Prompt template for Vicuna.
    }
   \end{table}
\newpage
\subsection{Prompt Template for Tulu}
We use the following format for prompting Tulu.

\begin{table}[h]
    \centering
    \small
    \begin{tabular}{p{0.85\linewidth}}
        \toprule

            \vspace{-2mm}
        \textbf{\textsc{Instruction:}} \\ <|user|> \\ Read the passage and answer the question. You may do some analysis on the passage and question before answering. But you should directly give out in the answer in the end and in the form of {{answer}} \\  <|assistant|> OK, I got it. \\ <|user|> \\

        \midrule
               \textbf{\textsc{Examples:}}\\
        Passage: [[example\_passage\_1]] \\
        Question: [[example\_question\_1]] \\
        <|assistant|>\\ 
        Answer: [[example\_answer\_1]] \\
        ...\\
        <|user|> \\
        Passage: [[example\_passage\_K]] \\
        Question: [[example\_question\_K]] \\
        <|assistant|>\\
        Answer: [[example\_answer\_K]] \\
        \midrule
        \textbf{\textsc{Example K+1:}}  \\
        <|user|> \\
        Passage: [[passage]] \\
        Question: [[question]] \\
        <|assistant|>\\
        \bottomrule
    \end{tabular}
        \caption{
  Prompt template for Tulu.
    }
   \end{table}

\newpage
\subsection{Prompt Template for GPT-NeoXT}

GPT-NeoXT is also allowed with specific format, as shown below.

\begin{table}[h]
    \centering
    \small
    \begin{tabular}{p{0.85\linewidth}}
        \toprule

            \vspace{-2mm}
        \textbf{\textsc{Instruction:}} \\ 
        ***Read the passage and answer the question.*** \\
        ***You may do some analysis on the passage and question before answering.*** \\
        ***But you should directly give out in the answer in the end and in the form of {{answer}}.*** \\

        \midrule
               \textbf{\textsc{Examples:}}\\
        <human> \\
        Passage: [[example\_passage\_1]] \\
        Question: [[example\_question\_1]] \\
        <bot>: [[example\_answer\_1]] \\
        ... \\
        <human> \\
        Passage: [[example\_passage\_K]] \\
        Question: [[example\_question\_K]] \\
        <bot>: [[example\_answer\_K]] \\
        \midrule
        \textbf{\textsc{Example K+1:}}  \\
        <human> \\
        Passage: [[passage]] \\
        Question: [[question]] \\
        <bot>: \\
        \bottomrule
    \end{tabular}
        \caption{
  Prompt template for Chat-NeoXT.
    }
   \end{table}

\clearpage
\section{Dataset Diversity Analysis}

\begin{figure}[h]
    \centering
    \includegraphics[width=0.48\textwidth]{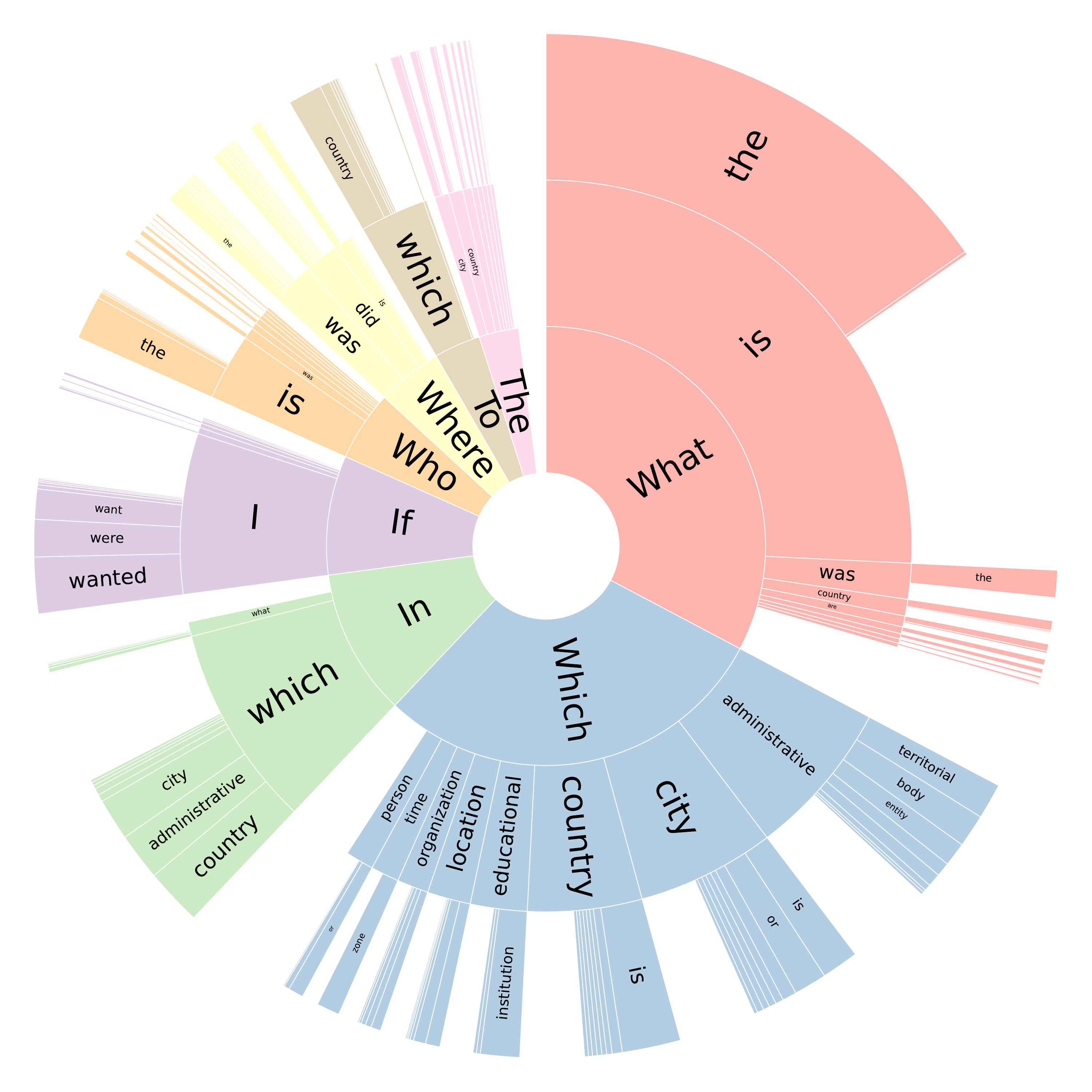}
    \caption{Trigram distribution of questions in \dataname.}
    \label{fig:question-trigram}
\end{figure}

One natural question arise from the construction process of \dataname is that, does \texttt{text-davinci-003} generate high quality questions?
We examine the quality by their diversity and plot the trigrams in Figure~\ref{fig:question-trigram}.
As we can see, although the questions are generated automatically, they are potentially different with each other.

\subsection{Question Diversity}

\newpage
\subsection{Rationale Diversity}
\label{app:retionale_diversity}

\begin{figure}[h]
    \centering
    \includegraphics[width=0.48\textwidth]{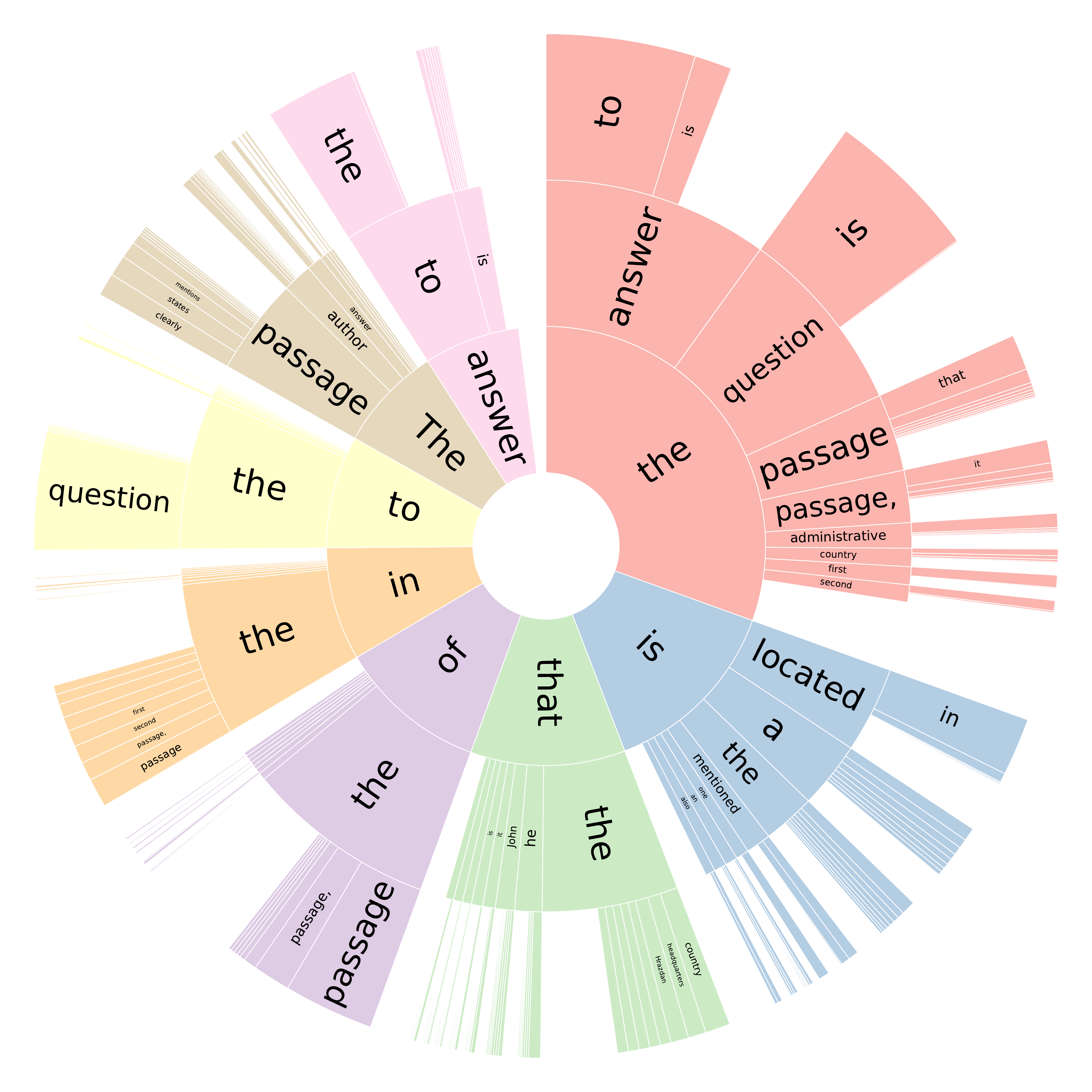}
    \caption{Trigram distribution of the annotated rationale.}
    \label{fig:rationale-trigram}
\end{figure}

The rationales are generated from human annotations for LLMs fine-tuning.
We plot the trigram of the rationales in Figure~\ref{fig:rationale-trigram}, which verifies that our human annotators provides diverse rationales.

\clearpage
\section{Rationale Case Study}
\label{app:rationale_case}

We show a rationale labeled by our annotators in Table~\ref{tab:rationale}.

\begin{table}[h]
    \centering
    \small
    \begin{tabular}{p{0.85\linewidth}}
        \toprule

            \vspace{-2mm}
        \textbf{\textsc{Passage:}} \\ 
        The London Borough of Waltham Forest is an administrative territorial entity located in Greater London. It is situated in the UTC+01:00 time zone and shares borders with the London Borough of Enfield. Waltham Forest is home to the capital city of Walthamstow, a vibrant area with a rich culture and history. It is the birthplace of two renowned writers, Eliza Parsons and Tom Hood, and the place of passing of writer Jane Ray. Additionally, the borough also contains the remote hamlet of Niaqornat. \\
        Waltham Forest has a lot to offer in terms of its culture and history. It has a variety of museums, parks, schools and other cultural venues. It also has a number of historic buildings, such as the Grade I listed St Mary's Church. The borough is also home to a number of landmarks, such as the William Morris Gallery and Epping Forest. It is also home to a number of festivals and events, such as the annual Walthamstow Garden Party. \\
        The London Borough of Waltham Forest is an area of great diversity and culture, and it is easy to see why so many people choose to make it their home. With its rich history and vibrant culture, it is a great place to live, work, and explore. \\

        \midrule
               \textbf{\textsc{Question:}}\\
        In which country is Niaqornat located? \\

        \midrule
        \textbf{\textsc{Rationale:}}  \\
        We can get the infomation that Niaqornat is located in the London Borough of Waltham Forest from the passage from the passage. And we all know that the London Borough of Waltham Forest is located in the United Kingdom. So we can infer that Niaqornat is located in the {{United Kingdom}}. \\
        \bottomrule
    \end{tabular}
        \caption{
  One sampled rationale for \datanameI.
    }
    \label{tab:rationale}
\end{table}

\newpage
\section{Evaluation Metric Analysis}
\label{app:eval_analysis}
\begin{figure}[h]
    \centering
    \includegraphics[width=0.95\linewidth]{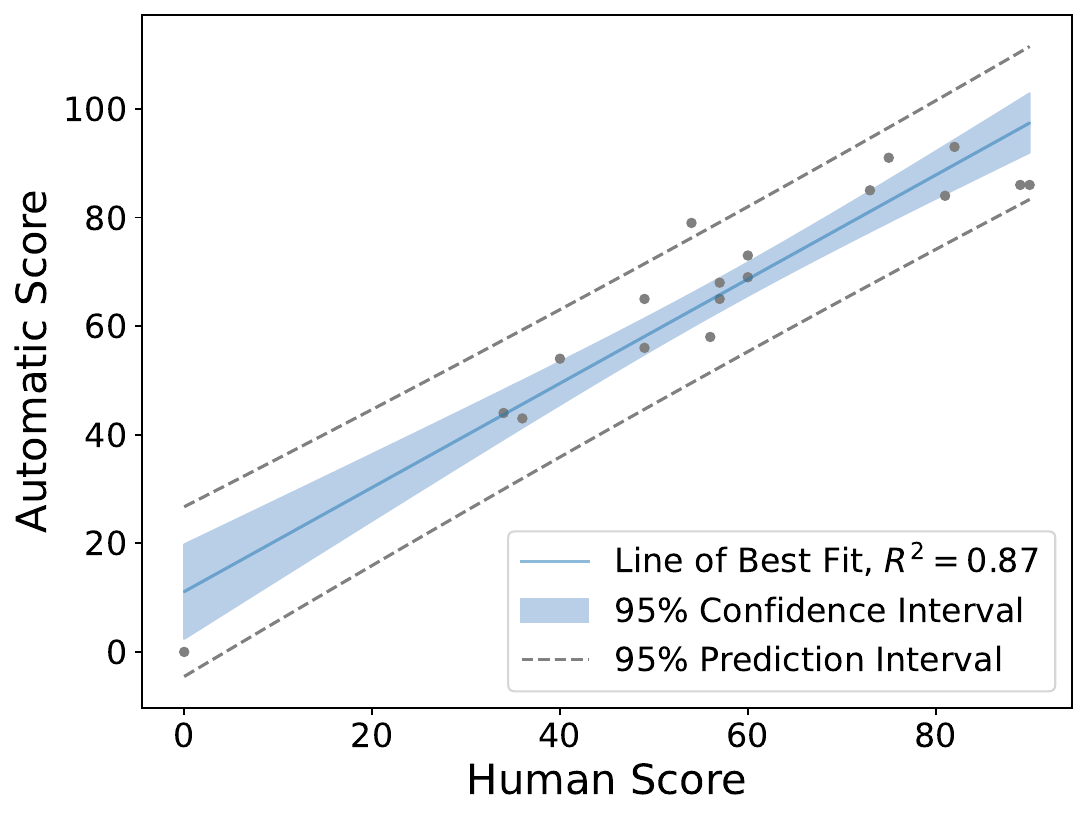}
    \caption{A scatter plot comparing human-annotated scores and automatic scores for $100$ randomly sampled answers from each model in \dataname}
    \label{fig:human-correlation}
\end{figure}

In order to investigate the feasibility of automatic evaluate strategy we proposed in section~\ref{sec:metric}, we randomly sample out $100$ answers of each model in \dataname, asking human annotator to mark the correctness of model answers. Figure~\ref{fig:human-correlation} shows the scatter plots of human annotated score verse the automatic score, which presents the highly positive linear correlation between two scores, suggesting that the automatic evaluation strategy is plausible when compare the performance between LLMs.

\newpage
\section{Effect of Example Number}
\begin{figure}[h]
    \centering
    \includegraphics[width=0.48\textwidth]{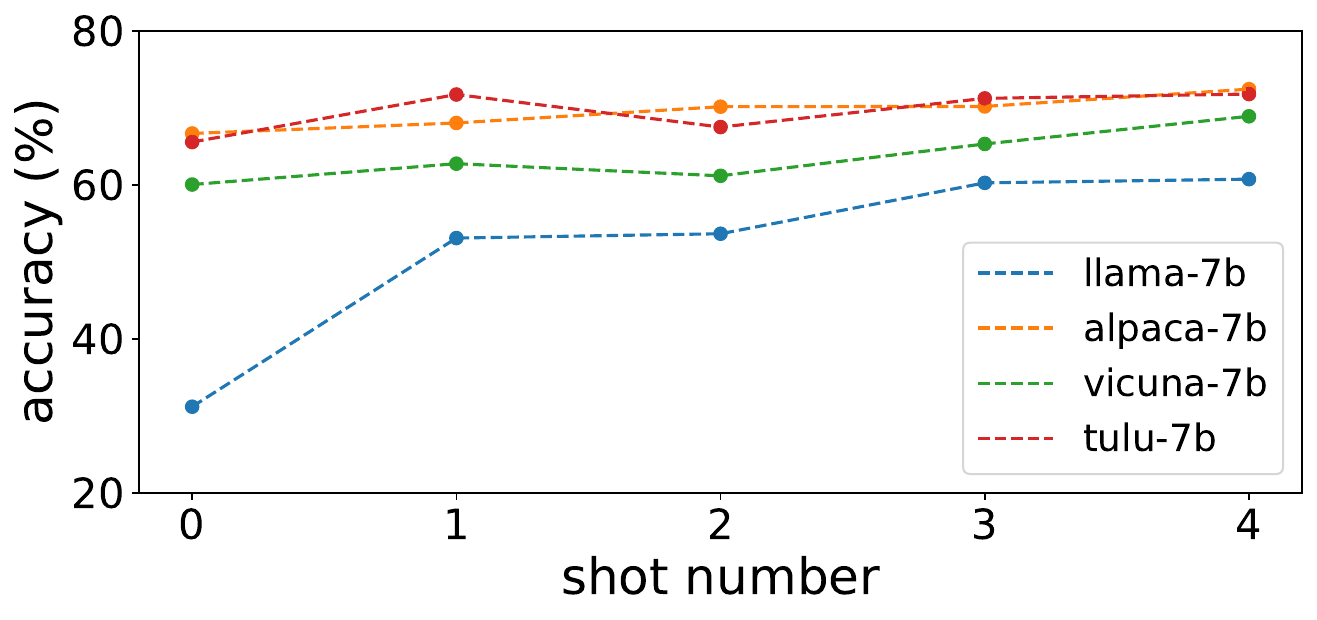}
    \caption{Accuracy (\%) with different example numbers.}
    \label{fig:example-number}
\end{figure}

In table~\ref{tab:main-result} and table~\ref{tab:prompting}, we report the performance of different LLMs with different prompting strategies under one-shot setting.
To thoroughly investigate the effect of example number, we test the performance of different LLMs from zero-shot to four-shot settings. Figure~\ref{fig:example-number} shows the results.

These results are averaged over KNOT-E, KNOT-S, and KNOT-I, with dataset sizes of 3887, 1136, and 510 respectively serving as weights. As observed, performance does improve with an increase in the number of shots, but this improvement tends to converge when the number of shots exceeds two. Additionally, due to the text window size limit (2048 max), the number of shots is restricted to four.

\end{document}